\documentclass{article}
\usepackage{spconf,amsmath,graphicx}
\usepackage{booktabs}

\usepackage{xcolor}

\title{Knowledge distillation for improved accuracy in spoken question answering}

\name{Chenyu You$^{1 \dagger}$\thanks{$^{\dagger}$ Indicates equal contribution.} \qquad 
Nuo Chen$^{2 \dagger}$ \qquad Yuexian Zou$^{2,3,*}$ \thanks{ Special acknowledgements are given to AOTO-PKUSZ JointResearch Center for Artificial Intelligence on Scene Cognition Technology Innovation for its support. $^{*}$ Corresponding Author.}}
\address{$^{1}$Department of Electrical Engineering, Yale University, CT, USA\\
$^{2}$ADSPLAB, School of ECE, Peking University, Shenzhen, China\\
$^3$Peng Cheng Laboratory, Shenzhen, China\\
\ninept \texttt{chenyu.you@yale.edu, \{nuochen,zouyx\}@pku.edu.cn}
}

\begin{document}
%
\maketitle
\begin{abstract}
Spoken question answering (SQA) is a challenging task that requires the machine to fully understand the complex spoken documents. Automatic speech recognition (ASR) plays a significant role in the development of QA systems. However, the recent work shows that ASR systems generate highly noisy transcripts, which critically limit the capability of machine comprehension on the SQA task. To address the issue, we present a novel distillation framework. Specifically, we devise a training strategy to perform knowledge distillation (KD) from spoken documents and written counterparts. Our work aims at distilling rich knowledge from the language model to improve the performance of the student model by reducing the misalignment between automatic and manual transcripts. Experiments demonstrate that our approach outperforms several state-of-the-art language models on the Spoken-SQuAD dataset.

\end{abstract}
\section{Introduction}
\label{sec:intro}
Text-based question answering (QA) \cite{huang2018flowqa,zhu2018sdnet,chen2020adaptive} is an important task in natural language processing (NLP), which requires the machine  to find the relevant textual answers in a context given a natural language question. In recent years, deep neural networks have achieved substantial progress in various QA applications. However, the voice interfaces for QA systems are less investigated. The main reason is the lack of large datasets with abundant annotations. Spoken-SQuAD \cite{li2019exploiting} is one of the few available datasets for spoken question answering (SQA) tasks, which leverages the text documents to generate corresponding spoken documents via Google Text-to-Speech system. Meanwhile, different from traditional text-based QA tasks, spoken question answering (SQA) includes audio signal processing, passage comprehension, and contextual understanding. Previous SQA approaches first transfer spoken content into text transcripts via ASR systems, then deploy some effective methods such as similarity matching \cite{DBLP:journals/corr/abs-1910-04958}, information retrieval \cite{fan2020spoken} to predict answers given the ASR transcriptions. Previous work has shown that ASR errors (e.g., Barcelona” to “bars alone") make the transcribed content much more difficult for the QA systems to understand and make reasonable predictions.

\begin{figure}
    \centering
    \includegraphics[width=0.9\linewidth]{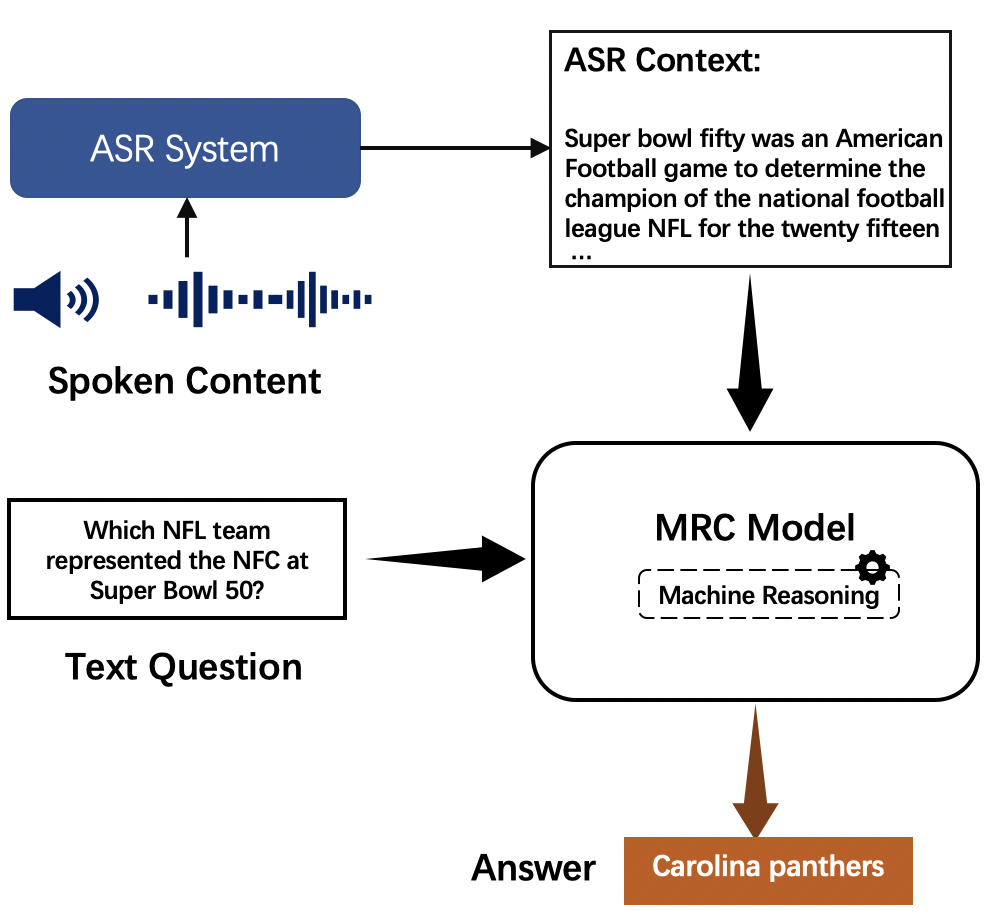}
    \caption{Flow diagram of SQA systems on Spoken-SQuAD dataset.}
    \label{fig:fig1}
    \vspace{-10pt}
\end{figure}


 
The combination of ASR errors and loose syntax are the major impediments when handling spoken documents. To mitigate the impact of transcription errors, several methods~\cite{li2018spoken,lee2018odsqa,you2020contextualized,DBLP:conf/icassp/LeeCL19,su2020improving,you2020data} have been proposed. Some recent methods \cite{li2018spoken,lee2018odsqa} adopt sub-word unit features strategy to enhance the word embedding with sub-word embedding to reduce unexpected ASR errors. Since ASR errors can be regarded as substitutions of word sequences for other ones (e.g., "feature" to "feather"), some sub-word sequences can still be correctly recognized when the corresponding transcribed content is wrong. However, it is still unclear whether we can explicitly learn relations in the discourse for the downstream tasks such as building QA systems. More recently, Lee et al. \cite{DBLP:conf/icassp/LeeCL19} proposes an adversarial domain adaptation method to reduce the gap between ASR hypotheses and the corresponding written content. Although the previous study has shown promising performance gains, it may suffer instability in training.


In this paper, we focus on the spoken question answering tasks. To resolve the challenges, we present a simple yet effective distillation approach, which fuses ASR-robust features to reduce the misalignment between ASR hypotheses and the reference transcripts. More concretely, we first take advantage of the dual nature property between speech utterances and text content by training the~\textit{teacher} model to learn two forms of the correspondences. We then distill this knowledge to the~\textit{student} model to mitigate the effect of unexpected transcript errors to boost the SQA performance.

\section{Method}
\subsection{Task definition}
Given a dataset $\mathcal{X} \in \{Q_i,D_i,A_i\}_{i}^{N}$, where $Q_i$ is a question in text or spoken form, $D_i$ is a document (passage) in spoken form and $A_i$ is the answer in text form, respectively. Specifically, $Q_i$ and $A_i$ are both in the form of a single sentence, while $D_i$ consists of multiple sentences. In this study, we use the Spoken-SQuAD dataset to validate the effectiveness of our approach on the extractive SQA task. Our goal is to find the text-based answer span $A_i$ from the transcriptions of spoken document $D_i$ given the question $Q_i$.



In this work, our QA system for spoken documents includes two sub-modules: ASR module and question answering module. More concretely, the QA module predicts the answer given a written question and speech transcriptions. We present the flow diagram of SQA on Spoken-SQuAD dataset in Figure \ref{fig:fig1}.

\begin{figure*}[t]
    \centering
    \includegraphics[width=0.9\linewidth]{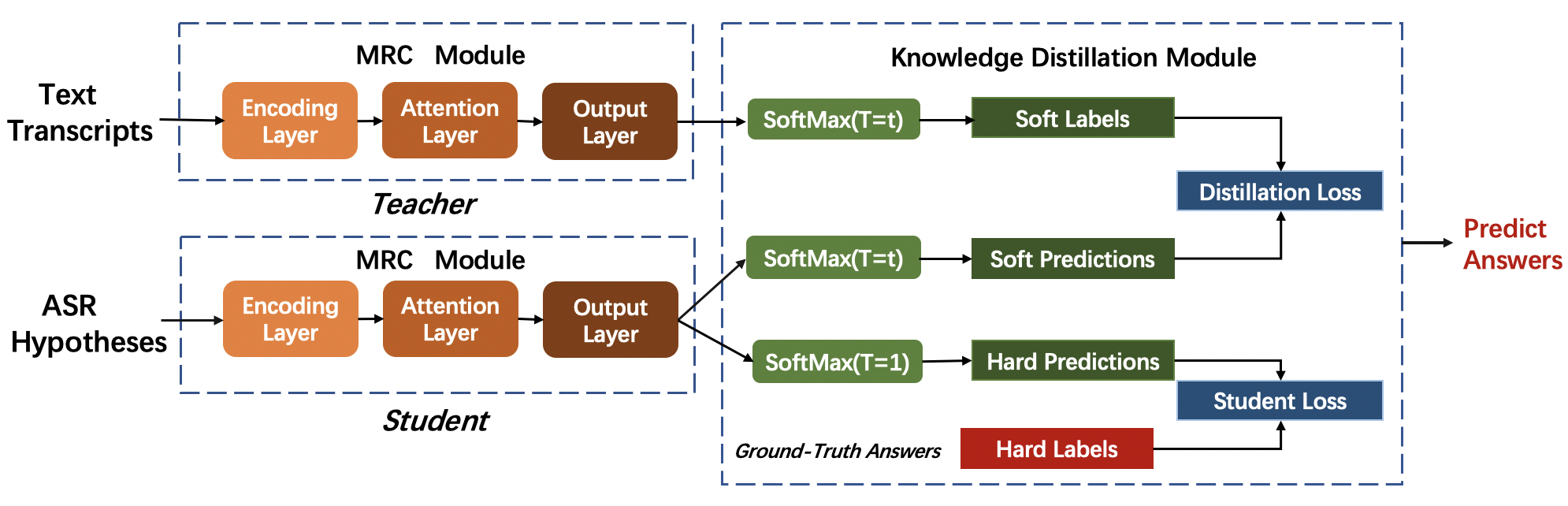}
    \caption{An overview of our proposed framework.}
    \label{fig:i}
    \vspace{-10pt}
\end{figure*}

\subsection{Knowledge Distillation}
Hinton et al.~\cite{hinton2015distilling} first introduced the idea of~\textit{knowledge distillation} (KD). In recent years, KD-based methods have achieved remarkable success in many computer vision and natural language processing tasks, such as natural language inference~\cite{jiao2019tinybert},sentiment classification~\cite{li2019exploiting}, and object detection~\cite{chen2017learning}. The key idea behind KD is to apply the prediction output of the~\textit{teacher} model as a ‘soft label’, and then utilize it as a supervision signal to guide the training of the~\textit{student} model. We set the model trained on the text transcripts as the~\textit{teacher} model $F_{T}(\cdot)$ and trained on the ASR hypotheses as the~\textit{student} model $F_{S}(\cdot)$, respectively.

Specifically, let $x$ and $y$ be the training input and ground-truth label from the dataset $\mathcal{X}$. Suppose the $z_{S}$ and $z_{T}$ are the output logits by $F_{S}(\cdot)$ and $F_{T}(\cdot)$, $F_{S}(\cdot)$ is trained to optimize the following objective:
\begin{equation*}
    L = \sum_{x\in \mathcal{X}} (\alpha \tau^2 \mathcal{KL}(p_{\tau}(z_{S}), p_{\tau}(z_{T})) + (1 - \alpha) \mathcal{XE}(z_{T}, y) ),
\end{equation*}
where $\mathcal{KL}(\cdot)$ and $\mathcal{XE}(\cdot)$ represent the Kullback-Leibler divergence and cross entropy, respectively. $p_{\tau}(\cdot)$ denotes the softmax function with temperature $\tau$, and $\alpha$ is a balancing factor.

\begin{table*}[t!]
\footnotesize
    \caption{Comparison of four baselines~(BiDAF, QANet, BERT, ALBERT). Note that, for brevity, we refer Text-SQuAD dev set and Spoken-SQuAD test set as T-SQuAD  dev and S-SQuAD test, respectively.}
    \centering
    \begin{tabular}{l | c c|c c|c c|c c}
        &\multicolumn{4}{c}{\textbf{T-SQuAD}}&
        \multicolumn{4}{c}{\textbf{S-SQuAD}}\\
        \hline\hline
        &\multicolumn{2}{c}{\textbf{T-SQuAD dev}}&
        \multicolumn{2}{|c}{\textbf{S-SQuAD test}}&\multicolumn{2}{|c|}{\textbf{T-SQuAD dev}}&
        \multicolumn{2}{|c}{\textbf{S-SQuAD test}}\\
        \textbf{Methods} &EM &F1 &EM &F1 & EM&F1 &EM&F1 \\
        \hline\hline
        BiDAF \cite{DBLP:journals/corr/SeoKFH16} & 65.8& 74.1 & 42.4& 54.8 & 44.9& 56.6 & 46.1& 58.9  \\
        QANet \cite{DBLP:journals/corr/abs-1804-09541} & 70.9 & 80.5  & 49.5 & 63.2 & 51.1 & 65.5  & 54.5 & 66.1\\
        BERT-large~\cite{DBLP:conf/naacl/DevlinCLT19} & 83.7 & 88.7  & 60.8 & 72.3 & 58.3 & 70.2  & 58.6 & 71.1\\
        ALBERT-large~\cite{lan2019albert}  & 84.4& 89.6 &61.6& 73.8 & 59.1&71.9 &59.4& 72.2\\
        \hline
        Average  & 76.2 &83.2 &53.6& 65.3 & 53.4 & 66.1 & 54.6 & 67.1\\
    \end{tabular}
    \label{tab:my_label_1}
    \vspace{-10pt}
\end{table*}

\subsection{Models}
To demonstrate the effectiveness and generality of our method, we choose two state-of-the-art QA models (BiDAF~\cite{DBLP:journals/corr/SeoKFH16}, QANet~\cite{DBLP:journals/corr/abs-1804-09541}, and two pre-trained language models: BERT~\cite{DBLP:conf/naacl/DevlinCLT19} and ALBERT~\cite{lan2019albert}) as the baseline, which achieve superior performance on Text-SQuAD~\cite{rajpurkar-etal-2016-squad} dataset.

\section{Experiments}
\subsection{Dataset}
Spoken-SQuAD (S-SQuAD)~\cite{li2018spoken} is an English listening comprehension dataset, which contains 37,111 question pairs in the training set and 5,351 in the testing set. In Spoken-SQuAD, the documents are in spoken form, and the questions and answers are in the text form. The word error rate (WER) is 22.77$\%$ on the training set and 22.73$\%$ on the testing set. The reference written documents in Spoken-SQuAD are from Text-SQuAD \cite{rajpurkar-etal-2016-squad}, which is one of the most popular machine reading comprehension datasets. Text-SQuAD (T-SQuAD)~includes 536 articles randomly selected from Wikipedia and the corresponding human-made 107,785 question-answer pairs.

In our experiments, we set our baselines trained on the Text-SQuAD as the \textit{teacher} model and trained on the Spoken-SQuAD as the \textit{student} model. Specially, we first train baselines on the Text-SQuAD training set and evaluate the performances of testing baselines on Text-SQuAD dev set and Spoken-SQuAD test set, respectively. Then, we train the baselines on the Spoken-SQuAD training set and evaluate the baselines on the Text-SQuAD dev set and Spoken-SQuAD test set, respectively. We report the quantitative results in Table \ref{tab:my_label_1}.


\subsection{Implementation Details}
In our experiments, we use PTB tokenizer and spaCy tokenizer for BiDAF and QANet, respectively. For BiDAF and QANet, we adopt pre-trained word vectors, GloVe~\cite{DBLP:conf/emnlp/PenningtonSM14} word vectors to achieve the fixed word embedding of each word from both Text-SQuAD and Spoken-SQuAD. The batch size is set to be 10. We set the learning rate as 1e-3. As for two pre-trained models (BERT and ALBERT), we use BPE tokenizer. We use a batch size of 2. The learning rate is set to be 2e-5. We utilize the large version of BERT and ALBERT in our experiments, which composes of 24 transformer layers, and the hidden size of the word vector is 1024 in this study. We train four baselines (BiDAF, QANet, BERT, ALBERT) for 20, 15, 3, 3 epochs, respectively. To maintain the integrity of all evaluated model performance, we adopt standard implementations and hyper-parameters of four baselines for training. We set the balancing factor $\alpha$ as 0.9, and the temperature $\tau$ is 2. We use the F1 and Exact Match (EM) to evaluate SQA model performance.

\begin{figure*}[t]
    \centering
    \includegraphics[scale=0.6]{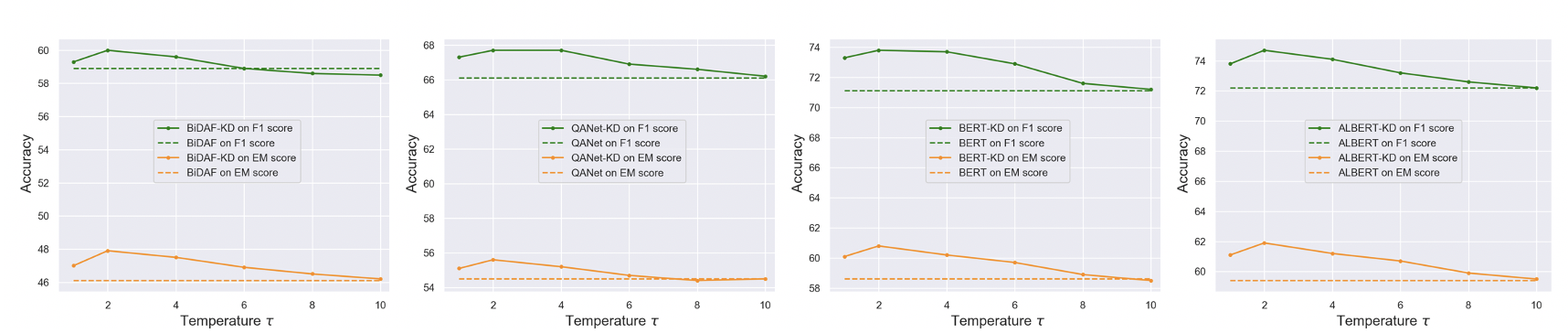}
    \caption{Ablation studies of temperature $\tau$ on four baseline performances~(BiDAF, QANet, BERT, ALBERT). Green and Orange denote the results of F1 score and EM score on  Spoken-SQuAD test set, respectively.}
    \label{fig:my_label}
    \vspace{-10pt}
\end{figure*}

\section{Results}
To investigate the effects of ASR error on the SQA task, we compare four baselines. Table~\ref{tab:my_label_1} reports the quantitative results. Table~\ref{tab:my_label_1} shows that, in all baseline, the performance trained on ASR transcriptions is far lower than that of training on the text documents. This suggests that the word recognition errors affect significant content words, which lead the SQA model to make predictions incorrectly. Taking ALBERT-large as an example, when training on the T-SQuAD and testing on the T-SQuAD, it achieves 84.4$\%$/89.6$\%$ on EM/F1 score. In contrast, when training on the S-SQuAD and testing on the S-SQuAD test set, ALBERT-large only obtains 59.4$\%$/72.2$\%$ on EM/F1 score. This confirms that ASR errors severely degrade the performance of SQA models. Thus, it is necessary to explore how to mitigate the impact of ASR errors for SQA systems.






To demonstrate the effectiveness of our proposed method, we compare four baseline methods~\textit{with} and~\textit{without} KD. We adopt the following setting - the~\textit{student} and~\textit{teacher} enjoy exactly the same network architecture. As shown in Table~\ref{tab:my_label_3}, it demonstrates that adopting the~\textit{knowledge distillation} strategy consistently boost the remarkable performance on all baselines. For BiDAF, our method achieves 46.5$\%$/58.7$\%$ (vs.44.9$\%$/56.6$\%$) and 47.9$\%$/60.0$\%$ (vs.46.1$\%$/58.9$\%$) in terms of EM/F1 score over the text documents and ASR transcriptions, respectively. For QANet, our method outperforms the baseline without distillation, achieving 52.7$\%$/67.9$\%$ (vs.51.1$\%$/65.5$\%$) and 55.6$\%$/67.7$\%$ (vs.54.5$\%$/66.1$\%$) in terms of EM/F1 score. As for two BERT-based models (BRET-large and ALBERT-large), our methods with KD consistently improve EM/F1 scores to 60.1$\%$/72.2$\%$ (vs.58.3$\%$/70.2$\%$) and
58.6$\%$/71.1$\%$ (vs.60.8$\%$/73.8$\%$);
60.8
$\%$/73.6$\%$ (vs.59.1$\%$/71.9$\%$) and 61.9$\%$/74.7$\%$ (vs.59.4$\%$/72.2
$\%$), respectively. These results confirm the importance of~\textit{knowledge distillation} in this setting.



\begin{table}
    \caption{Comparison of our method. We set the model on  text corpus as the~\textit{teacher} model, and the one on the ASR transcripts as the~\textit{student} model.}
    \centering
    \footnotesize
    \begin{tabular}{l|cc|cc}
    \toprule
    & \multicolumn{2}{|c|}{\textbf{T-SQuAD dev}}&\multicolumn{2}{|c}{\textbf{S-SQuAD test}}\\
    \textbf{Methods}&EM&F1 &EM&F1\\
    \hline\hline
   BiDAF \cite{DBLP:journals/corr/SeoKFH16}&44.9&56.6&46.1&58.9\\
   \quad+ Sub-word \cite{li2018spoken}&45.8&57.7&47.1&59.4\\
   \quad+ Do-Adaptation \cite{DBLP:conf/icassp/LeeCL19}&45.9&57.4&47.0&59.1\\
    \quad+ \textbf{KD} &\textbf{46.5}&\textbf{58.7}&\textbf{47.9}&\textbf{60.0}\\
    \hline
    QANet \cite{DBLP:journals/corr/abs-1804-09541}&51.1&65.5&54.5&66.1\\
    \quad+ Sub-word \cite{li2018spoken}&51.9&66.6&55.0&66.6\\
    \quad+ Do-Adaptation \cite{DBLP:conf/icassp/LeeCL19}&52.2&66.7&55.2&66.8\\
    \quad+ \textbf{KD}&\textbf{52.7}&\textbf{67.9}&\textbf{55.6}&\textbf{67.7}\\
    \hline
    BERT-base \cite{DBLP:conf/naacl/DevlinCLT19}&58.3&70.2&58.6&71.1\\
    \quad+ Sub-word \cite{li2018spoken}&59.3&71.1&59.7&72.4\\
    \quad+ Do-Adaptation \cite{DBLP:conf/icassp/LeeCL19}&59.5&71.2&59.8&72.6\\
    \quad+ \textbf{KD}&\textbf{60.1}& \textbf{72.2}& \textbf{60.8}& \textbf{73.8}\\
    \hline
    ALBERT-base \cite{lan2019albert}&59.1&71.9&59.4&72.2\\
    \quad+ Sub-word \cite{li2018spoken}&59.7&72.8&60.7&73.5\\
    \quad+ Do-Adaptation \cite{DBLP:conf/icassp/LeeCL19}&60.0&72.7&60.8&73.5\\
    \quad+ \textbf{KD}&\textbf{60.8}&\textbf{73.6}&\textbf{61.9}&\textbf{74.7}
    \\
    \bottomrule
    \end{tabular}
    \label{tab:my_label_3}
    \vspace{-10pt}
\end{table}

To analyze the effect of the pre-trained language model on the SQA task, we first pick SDNet~\cite{zhu2018sdnet}, which uses the pre-trained BERT-based and LSTM as its embedding encoder, and 283M parameters in total. For comparison, we establish the lightweight version of SDNet (17.7M parameters) by removing the pre-training language model (BERT-base) from SDNet. As shown in Table~\ref{tab:my_label}, we can observe that, on the Spoken-SQuAD, the original SDNet consistently outperforms the one without BERT. This suggests that the importance of the pre-trained language model on the SQA task. Meanwhile, to demonstrate the capability of knowledge distillation, we set the original SDNet trained on the Spoken-SQuAD as the massive \textit{teacher} model, and SDNet without BERT trained on the Spoken-SQuAD as the lightweight \textit{student} model. We find that performing~\textit{knowledge distillation} can indeed improve the performance of QA on speech transcripts. This suggests that distilling highly effective language representations learned by the large-scale~\textit{teacher} model to the small~\textit{student} model, which can improve the training process and the performance of the compact model.




\begin{table}[t]
\caption{Comparison the performance of SDNet \cite{zhu2018sdnet} and SDNet  \cite{zhu2018sdnet}without (w/o) BERT with repsect to manual transcriptions (Text-SQuAD) or ASR hypothesis (Spoken-SQuAD).  }
    \centering
    \footnotesize
    \begin{tabular}{l|cc|c|c}
    \toprule
    \textbf{Model}&\multicolumn{2}{c}{\textbf{Data Usage}}&
        \multicolumn{1}{c}{\textbf{Result}}&\textbf{Parameters}\\
        \hline
     &Training& Dev$\&$Test&EM/F1&Millions\\
     \hline\hline

    SDNet \cite{zhu2018sdnet} &Manual&Manual&78.6/87.1&283 \\
    w/o BERT&Manual&Manual&72.5/80.3&17.7\\
    \hline
    SDNet &ASR&ASR&57.8/71.8&241.4\\
    SDNet+\textbf{KD} &ASR&ASR&59.2/73.6&241.4 \\
    w/o BERT&ASR&ASR&51.8/63.7&15.1\\
     w/o BERT+\textbf{KD}&ASR&ASR&52.2/64.5&15.1\\
    \bottomrule
    \end{tabular}
    \label{tab:my_label}
    \vspace{-10pt}
\end{table}

\section{QUANTITATIVE ANALYSIS}
\textbf{Effect Of Temperature $\tau$.} \quad 
To investigate the impact of temperature $\tau$, we evaluate the performances of four baselines with the standard choice of the temperature~$\tau \in \{1,2,4,6,8,10\}$. As shown in Figure \ref{fig:my_label}, we find that four baselines consistently achieve their best performances in terms of F1 and EM scores with $T$ setting to 2. All models are trained on Spoken-CoQA dataset, and validated on the  Spoken-CoQA test set, respectively. It is noteworthy that we see  the similar trends are emerging when testing baselines on Text-SQuAD dev set. 


\noindent \textbf{Generalizability for Knowledge Distillation. }\quad 
Table~\ref{tab:my_label_3} shows that our proposed method significantly improves the network performance compared with the state-of-the-art baseline models on Spoken-SQuAD. As shown in Table \ref{tab:my_label}, we examine whether the~\textit{knowledge distillation} can mitigate the impact of ASR errors to achieve better accuracy in both cases over the ASR transcripts. There is some evidence that using KD strategy behaves better than the same model without KD. As can be seen, as the~\textit{teacher} model is much larger than the~\textit{student} model, the~\textit{student} is able to mimic the~\textit{teacher}, suggesting~\textit{teachers} yield better~\textit{students} accuracy than a~\textit{student} trained alone. In other words, it suggests that performing~\textit{knowledge distillation} is a promising research direction in the spoken question answering task.



\section{Conclusion}
In this paper, we have presented a data distillation approach to mitigate the impact of ASR errors. In our~\textit{teacher-student} framework, we distill the superior knowledge from the model trained on textual documents to the one on ASR transcripts to achieve better performance by taking advantage of interactivity in spoken-written documents to reduce transcript errors and loose syntax. Experiments demonstrate the effectiveness of our proposed model, suggesting the generalizability for knowledge distillation. We believe that our data distillation technique may be a promising direction to advance different speech processing and natural language processing tasks.



\bibliographystyle{IEEE.bst}
\bibliography{refs.bib}

\end{document}